\documentclass{scspaperproc}

\usepackage{latexsym}
\usepackage{graphicx}
\usepackage{mathptmx}

\usepackage{amsmath}
\usepackage{amsfonts}
\usepackage{amssymb}
\usepackage{amsbsy}
\usepackage{amsthm}

\usepackage[pdftex,colorlinks=true,urlcolor=blue,citecolor=black,anchorcolor=black,linkcolor=black,bookmarks=false]{hyperref}

\usepackage{hyphenat}
\newtheoremstyle{scsthe}% hnamei
{8pt}% hSpace abovei
{8pt}% hSpace belowi
{\it}% hBody fonti
{}% hIndent amounti1
{\bf}% hTheorem head fontbf
{.}% hPunctuation after theorem headi
{.5em}% hSpace after theorem headi2
{}% hTheorem head spec (can be left empty, meaning `normal')i

\theoremstyle{scsthe}

\begin{document}

%***************************************************************************
% AUTHOR: AUTHOR NAMES GO HERE
% FORMAT AUTHORS NAMES Like: Author1, Author2, and Author3 (last names)
%
%		You need to change the author listing below!
%               Please list ALL authors using last name only, separate by a comma except
%               for the last author, separate with "and"
%
%\SCSpagesetup{Krug, Raisch, Sigg, Aimer, Wirnsberger, Schäfer, and Tischler}

\SCSpagesetup{Krug, Raisch, Aimer, Wirnsberger, Sigg, Schäfer, and Tischler}

% AUTHOR: Uncomment ONE of these correct conference names.
\def\SCSconferencename{Annual Modeling and Simulation Conference}
%\def\SCSconferencename{Power Plant Simulation Conference}

% AUTHOR: Uncomment ONE of these correct conference acronyms.
\def\SCSconferenceacro{ANNSIM'25}
%\def\SCSconferenceacro{PowerPlantSim'24}

% AUTHOR: Set the correct year of the conference.
\def\SCSpublicationyear{2025}

% AUTHOR: Set the correct editors of the conference
\def\SCSconferenceeditors{J.L. Risco-Martín, G. Rabadi, D. Cetinkaya, R. Cárdenas, S. Ferrero-Losada, and A.A. Abdelnabi}

% AUTHOR: Set the correct month and dates; the dates are separated by a single minus sign
% with no spaces and no leading zeros, the month is a full name (e.g. April) with the first letter
% capitalized. For example, "April 8-13".
\def\SCSconferencedates{May 26-29}

% AUTHOR: Set the correct venue in the form "City, State, Country", for example, "Los Angeles, CA, USA".
\def\SCSconferencevenue{Universidad Complutense de Madrid, Madrid, Spain}

% AUTHOR: Enter the title, all letters in upper case
\title{BUILDA: A THERMAL BUILDING DATA GENERATION FRAMEWORK FOR TRANSFER LEARNING}

% AUTHOR: Enter the authors of the article
\author[\authorrefmark{1} \authorrefmark{2} \authorrefmark{3}]{Thomas Krug}
\author[\authorrefmark{1} \authorrefmark{2} \authorrefmark{4} ]{Fabian Raisch}
\author[\authorrefmark{2}]{Dominik Aimer}
\author[\authorrefmark{2}]{Markus Wirnsberger}
\author[\authorrefmark{2}]{\\Ferdinand Sigg}
\author[\authorrefmark{3}]{Benjamin Schäfer}
\author[\authorrefmark{2}]{Benjamin Tischler}

% Department and university with the mark
\affil[\authorrefmark{1}]{Both authors contributed equally to the research}
% Do NOT include emails in the title section. They will appear in the author biography

\affil[\authorrefmark{2}]{Technical University of Applied Sciences Rosenheim, Rosenheim, Germany}

\affil[\authorrefmark{3}]{Karlsruhe Institute of Technology, Karlsruhe, Germany}

\affil[\authorrefmark{4}]{Technical University of Munich, Munich, Germany}

\maketitle

\section*{Abstract}

Transfer learning (TL) can improve data-driven modeling of building thermal dynamics. Therefore, many new TL research areas emerge in the field, such as selecting the right source model for TL. However, these research directions require massive amounts of thermal building data which is lacking presently. Neither public datasets nor existing data generators meet the needs of TL research in terms of data quality and quantity.
Moreover, existing data generation approaches typically require expert knowledge in building simulation. 
We present BuilDa, a thermal building data generation framework for producing synthetic data of adequate quality and quantity for TL research. The framework does not require profound building simulation knowledge to generate large volumes of data. 
BuilDa uses a single-zone Modelica model that is exported as a Functional Mock-up Unit (FMU) and simulated in Python. 
We demonstrate BuilDa by generating data and utilizing it for pretraining and fine-tuning TL models.

\textbf{Keywords:} synthetic data generation, transfer learning, building thermal dynamics, data-driven modeling

%% AUTHOR:
% This is a list of no more than five keywords that will identify your paper in indices and databases (required).
% Do not use the words “computer”, “simulation”, “model”, or “modeling”, since these are all assumed.

\section{Introduction}
\label{sec:intro}

In 2021, building operations accounted for 27\% of global CO$_2$ emissions and 30\% of the world's energy demand \cite{globalstatus}. 
Recently, researchers have invested significant effort in developing data-driven models as they can serve as a basis for various tasks in intelligent building energy management. Many state-of-the-art approaches in building energy control or fault detection \& diagnosis (FDD) rely on a data-driven model that represents the thermal dynamics of the building \cite{YeYao2021MPCReview, Wang2019DDMTBD, chen2023reviewFDD}. However, such a model requires a sufficiently large amount of training data to achieve the desired quality. Depending on the task, acquiring measurement data over several months up to multiple years is often necessary. This is not feasible for most buildings, especially not for recently built ones.

A currently discussed solution to improve data efficiency, precision, and robustness of data-driven models is transfer learning (TL) \cite{pinto2022transfer, peirelinck2022transfer}. In TL, a model is pretrained with data from a source building to acquire knowledge. Subsequently, the pretrained model is fine-tuned for a target building using a relatively small amount of measured data specific to the target building. TL has already been successfully applied in various studies in the field of building dynamics \cite{chen2020transfer, jiang2019deep, pinto2022transfer}. However, TL requires source data in sufficient quantity and quality to achieve good results \cite{pinto2022transfer, jiang2019deep}. Recent developing research areas, such as research in generalized TL models \cite{2025gentl} or studies on building similarity (e.g. weather conditions, building size, occupancy patterns) for TL \cite{pinto2022sharing, li2024building,2024Similarity} underline the need for massive amounts of high-variance data with detailed and accessible metadata. Thorough studies on how specific building features influence the effectiveness of TL approaches would greatly enhance the understanding of the conditions under which TL is most effective.

Ideally, data for TL research purposes should exhibit a high variability to cover a wide range of possible building dynamics, representing numerous buildings and their environmental conditions. At the same time, the data source should provide detailed metadata descriptions. Yet, data with desired properties is rare, restricting TL research for building thermal dynamics. 
Currently, researchers either need to rely on publicly available datasets or they need to generate synthetic data by themselves. Both options have their own limitations, with the latter often requiring extensive knowledge of building simulation. Researchers without access to such expertise, including machine learning researchers, face significant challenges due to the lack of accessible and detailed data sources.   

\section{Background}
\label{sec:background}
Publicly available building datasets, whether real-world or synthetic \cite{miller2020building, miller2017building, li2021synthetic}, are typically static, offering little flexibility for customization. As the data is already generated or measured, the datasets cover only a fixed distribution of buildings and an in-depth exploration of the thermal dynamic behavior of the buildings that generated the data is not possible. 
Moreover, public data sources often lack detailed metadata. In \cite{2024Similarity}, the authors investigate the influence of building similarity on successful TL for cooling load prediction, but couldn't consider features of the building envelope in detail, due to inaccessible metadata. Similarly, \cite{pinto2022sharing} relied on a dataset with only three building efficiency levels, making granular parameter-level analysis difficult. In \cite{li2024building}, researchers had to manually infer weather conditions from the nearest weather station, highlighting data aggregation challenges.

An alternative is synthetic data generation using physics-based building models, e.g., modeled with building modeling libraries such as Modelica \cite{mattsson1997modelica} or EnergyPlus \cite{2001EnergyPlus}. While these models theoretically enable unlimited data generation, they require extensive domain expertise and are time-intensive to set up. Additionally, they are often not easily transferable across different building types, limiting their adaptability and usefulness for TL research.
Tools like eppy \cite{eppy} allow systematic modifications of EnergyPlus models but still demand considerable simulation expertise.

Another approach involves data generators, which integrate building modeling knowledge, making them more accessible to non-experts. 
The range of available data generators spreads from data augmentation frameworks such as \cite{Fan2022SynDataAug} to data generators for synthetic smart meter or electrical data \cite{Hong2020SynSmartMeter, charbonnier2024home}. However, only few focus on thermal building dynamics, such as Synconn\_build \cite{chaudhary2023synconn_build} or eplusr  \cite{eplusr}. Synconn\_build lacks flexibility in terms of adjustable building parameters, limiting its suitability for TL research. Eplusr, on the other hand, offers extensive configurability, but requires expertise in manipulating EnergyPlus IDF files. These challenges highlight the need for accessible, customizable, and scalable data-generation frameworks for TL research in building thermal dynamics, particularly for researchers without building simulation expertise.

We present BuilDa, a framework designed to generate high-fidelity thermal time-series data for TL without requiring simulation knowledge. BuilDa is highly customizable, supports parallelized data generation, and utilizes a validated building model. Users can adjust parameters such as weather, occupancy schedules, system controls, and building properties through a single configuration file. Additionally, we provide predefined profiles for climate, occupancy behavior, and control mechanisms.

The authors see the contribution of this work in the following aspects: (i) We provide an extensive data-generation framework for generating synthetic thermal multivariate time series building data, highly customizable, usable without expertise in building simulation and suitable for TL research. (ii) With the framework, a flexible, validated, high-fidelity single-zone building simulation model for data generation is delivered, incorporating extensive building modeling domain knowledge and covering a wide range of possible single-zone buildings. (iii) We generate synthetic data with BuilDa and pretrain and fine-tune a standard machine learning model to showcase the use for TL research by conducting a short study on the influence of building properties on the effectiveness of TL.

\section{Framework and Functionalities}
\label{cha:Framwork}

\subsection{Architecture}
\label{cha:Architecture}
This chapter provides an overview of the architecture of the BuilDa framework and its major building blocks. The BuilDa framework consists of two primary components. The first component is a physical Modelica building model (referred to as the base model) that simulates the thermal dynamics of a single-zone building. The second part is a framework to run multiple simulations with different parameters, implemented in Python. The framework runs the Modelica model as a Functional Mock-up Unit (FMU) via the FMPy library \cite{FMPy}, following the FMI standard 2.0.4 \cite{blochwitz2012functional} . Running the model as an FMU with Python means that no special knowledge of Modelica or additional tools like the BCVTB middleware \cite{BCVTB2008} are needed. 

\begin{figure}[htb]
    \centering
    \includegraphics[width=0.55\linewidth]{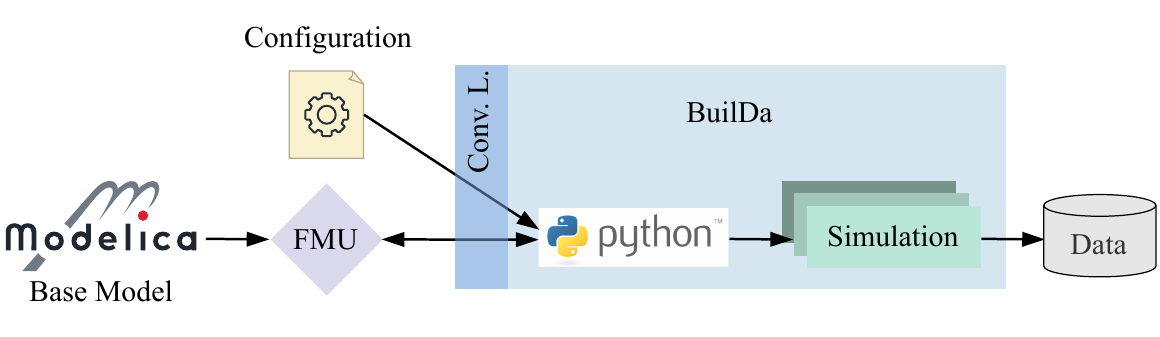}
    \caption{Architecture overview BuilDa.}
    \label{fig:Pipeline}
\end{figure}

\autoref{fig:Pipeline} illustrates the basic architecture of the BuilDa framework and the workflow to generate synthetic data. We provide a configuration file where users can set the parameters for the building model, control, as well as parameters of the simulation itself. We added a converter layer between the Modelica model parameters and the settings in the configuration file. With the converter layer, we simplify the user input, as some model parameters depend on each other. For example, the size of a room and its volume are directly related. In the configuration file, only one of these values needs to be set, while the dependent value is automatically calculated by the converter layer. \autoref{cha:Building Parameters} explains which parameters can be adjusted with BuilDa. The converter layer is specifically designed for our FMU. Thus, it may require modifications if another FMU with a different parameter set should be used.

\subsection{Building Model}
\label{cha:buildingmodel}
\begin{figure}[htb]
    \centering
    \includegraphics[width=0.43\linewidth]{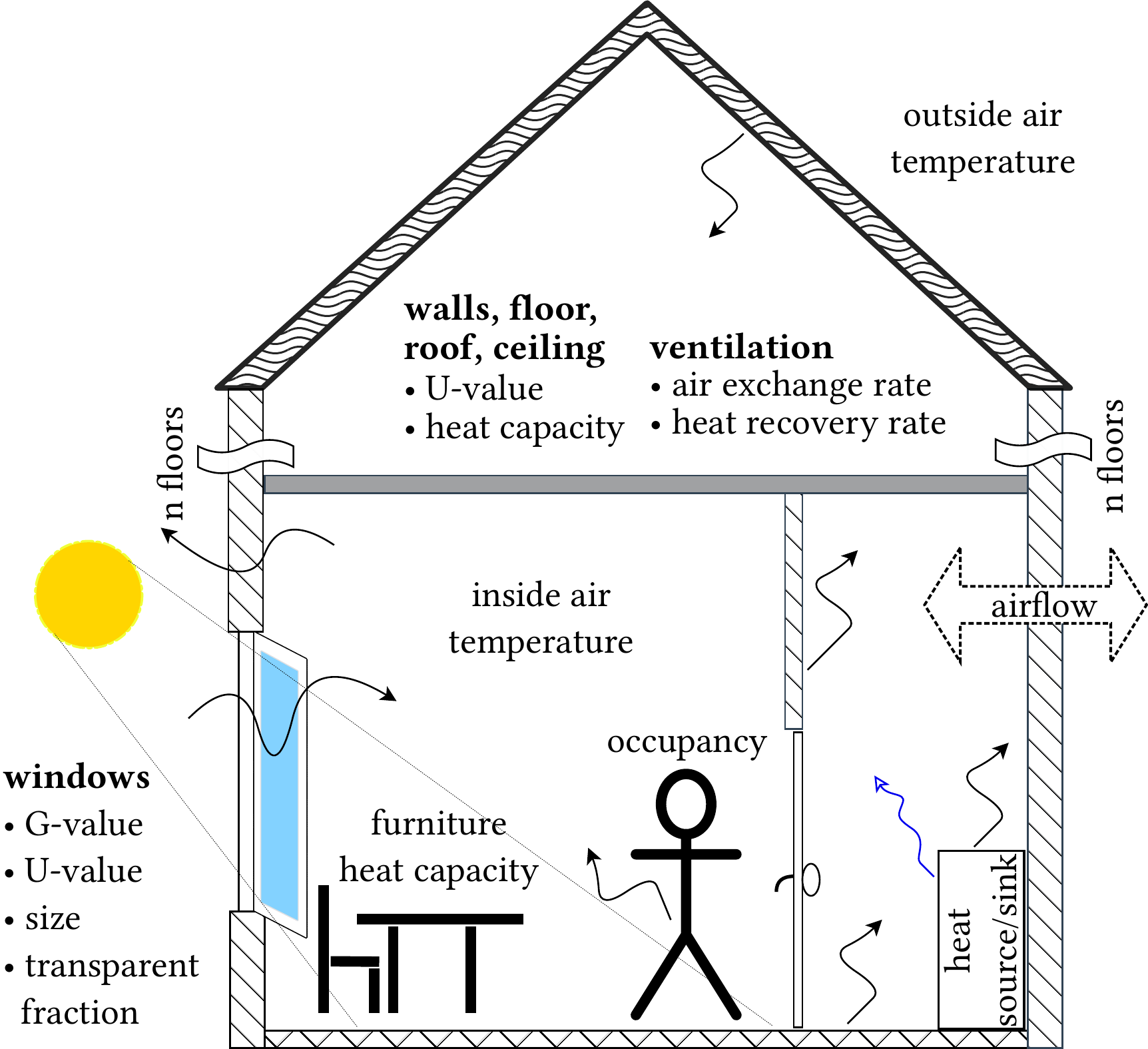}
    \caption{Basic setup of the heating system.}
    \label{fig:setup}
\end{figure}

The physical building model is based on the Modelica Buildings library 
\cite{Wetter014buildinglib}. To ensure accuracy, the building and its components are modeled based on the methods outlined in VDI6007 Part 1 for building simulations \cite{VDI6007-1}. For the building model, a uniform indoor air temperature is assumed, which applies to many types of buildings, especially residential ones. This simplification allows for easier analysis while still reflecting typical conditions found in most homes \cite{TASK44_2013}. \autoref{fig:setup} provides an overview of the single-zone building model, highlighting its key components and parameters. The heat flow into and out of the building are shown with arrows (blue arrow represents cooling). The model includes internal and external walls, roof, floor, ceilings, windows and furniture.  We used an ideal thermal source for cooling and heating. Ideal means it has a very low thermal mass, similar to an electric radiator, allowing heat to be transferred directly to the environment. The convective and radiative parts of the thermal source are set in the configuration file. The maximum heating power is calculated automatically based on DIN 18599 Part 2 by the converter layer (see \autoref{cha:Architecture}) \cite{DIN18599-2}. We also follow this standard for the calculation of the maximum cooling power, but assume simplifications. Solar gains and thermal losses are considered for the windows, exterior walls and roof. For the floor, only thermal losses are considered. Inside the zone, heat exchange between walls as well as between walls and the thermal source is modeled through long-wave radiation and convection. A summary of the modifiable parameters is shown in \autoref{tab:building_parameters_short}. In \autoref{cha:Building Parameters} and \autoref{cha:input_parameters}, we explain the parameters in more detail.

\subsubsection{Building parameters} 
\label{cha:Building Parameters}
 
Building parameters define the specific characteristic of a building, such as its size, envelope quality, thermal mass, windows, and ventilation. These characteristics influence the building’s thermal behavior. The user can set the \textbf{building size} by setting the length, width and height of one floor, as well as the number of floors. The building components, roof, floor, windows, interior and exterior walls, are modeled as R-C elements. The R-value represents thermal resistance, indicating the insulation \textbf{quality of the envelope}, while the C-value represents \textbf{thermal mass}. Three R-C nodes were chosen for discretization of the components. According to \cite{Davies2004}, this number provides sufficient accuracy for the model. However, the accuracy is influenced by the number and distribution of nodes, which depend on the wall structure. The heat capacity of the furniture is considered and added to the capacity of the internal walls. For the external walls and \textbf{windows}, orientation is considered to account for direction-dependent solar heat gains. Accordingly, the G-value of the windows is defined to represent the fraction of solar radiation transmitted through the glazing. 

For \textbf{ventilation}, heat flow (due to ventilation system, infiltration, or window opening) is determined by considering the air flows and the heat recovery rate (e.g., 0 for natural ventilation or exhaust-air systems, >0.7 for ventilation systems with heat recovery). An \textbf{openable window} is included in the model. Following the standard VDI 2078  \cite{VDI2078}, we have modeled the airflow through a fully opened window based on the indoor and outdoor temperatures, as well as the size and dimensions of the window. During the simulation, the window can be opened according to an external window opening profile (see \autoref{cha:input_parameters}) or by an external control signal (see \autoref{tab:building_parameters_short}), allowing the user to implement customized control logic.

\begin{table}[htb]
\small
\caption{Building simulation parameters.}
\label{tab:building_parameters_short}
\centering
    \begin{tabular}{p{1.3cm}|p{1.7cm}|p{11.8cm}} 
        \hline
\textbf{Category} & \textbf{Parameter}                          & \textbf{Explanation}                                                                                                                             \\ \hline
Building          & Building size                               & Thermal envelope size (including length, width, number of floors, and floor height)                                                              \\ \cline{2-3} 
                  & Quality \newline of envelope & Thermal quality of building envelope (U-value, structure)                                                                                        \\ \cline{2-3} 
                  & Thermal mass                                & Heat capacity of external and internal walls, roof, floor, furniture                                                     \\ \cline{2-3} 
                  & Windows                                     & Size, orientation, and quality of windows (U-value, G-value and transparent fraction)                                                            \\ \cline{2-3} 
                  & Ventilation                                 & Air exchange rates and heat recovery rate                                                                                                        \\ \hline
Input             & Controller                                  & Configurable external controller(s) (e.g., for heating, cooling, window opening) or internal controller configuration (setpoints)                                          \\ \cline{2-3} 
                  & Internal gains                                   & Internal gains profile, calculated from occupancy profile (optionally contains heat gains from electrical devices)                                                                                        \\ \cline{2-3} 
                  & Window opening                              & Window opening profile (influences ventilation heat losses), calculated from occupancy profile                                                                                           \\ \cline{2-3} 
                  & Weather                                     & Weather data with outside temperature, solar radiation, etc.\newline (main influence on heat energy demand and zone temperature) \\ 
\end{tabular}
\end{table}

\subsubsection{Input parameters} 
\label{cha:input_parameters}
 
Input parameters, in contrast to building parameters, represent parameters that configure input data for the model during the simulation, including controller, occupancy patterns, window opening behavior and weather conditions. BuilDa provides two types of \textbf{controllers}. The FMU has an internal controller for heating, which is proportional controller with night setback. The setpoints for day and night can be chosen individually. Alternatively, an external controller can be configured for heating, as well as for cooling and window opening, enabling more flexible and adaptive control strategies. To achieve this level of flexibility, we decoupled the frequency at which the controller updates the actuation signal from the rate at which the simulation retrieves its status and writes data.   

Occupancy in BuilDa affects \textbf{internal heat gains} and optionally \textbf{window opening} activity. To model this, a spreadsheet tool generates profiles for both internal heat gains (from people or electrical devices) and window opening activity. The tool creates a yearly occupancy profile, which is based on four day profiles (workday, Saturday, Sunday, holiday). Each profile has an individual schedule and may include additional internal heat gains, such as those from electrical devices. Window opening activity is updated every 5 minutes based on occupancy, sleep times, and ventilation awareness. The profiles can be exported in a CSV-like format for the FMU.

\textbf{Weather} significantly affects a building's thermal behavior. In BuilDa, users can simulate different locations by specifying the weather file path in the configuration file. The model imports weather data in MOS format. An open-source tool converts EPW weather files (available worldwide \cite{TMYx}) to MOS format. BuilDa also includes several weather files for central Europe.

\subsection{Software Usage}
\label{cha:TSUsage}

First, the parameter values need to be specified in the configuration file by the user (see appendix \autoref{tab:allParameters} for a list of the major parameters). BuilDa provides various ways to construct a set of variations of the values for simulation, e.g., setting a range of values with a step size or the construction of a cartesian product. BuilDa iterates through the set of variations and executes the simulations in parallel. The output is written to the file system as a CSV file, and output columns can be specified in the configuration file. The duration of a complete simulation run with BuilDa primarily depends on the size of the variation set, the controller's update sampling time, the FMU state output interval, and the number of available compute cores. For instance, simulating the FMU with 100 variations over one year, using an output interval of 300 seconds, required approximately 12.5 minutes (averaging 7.5 seconds per simulation) when using the internal controller. In contrast, the same simulation with an external two-point controller, updating the manipulated variable in 900-second steps, took approximately 48 minutes (averaging 29 seconds per simulation). The simulation was executed on an Intel® Core™ i7-6600U CPU (2.60 GHz, 4 cores) running Ubuntu 22.04 with 20 GB of RAM. The increased execution time in the externally controlled FMU is attributed mainly to additional recalculations within the FMU triggered by each controller update. Lastly, we want to note that the existing FMU in our framework can be replaced with a different one. We provide a description on how to perform the adaptation of our framework to a new FMU. The BuilDa framework is available on GitHub \cite{BuilDaRepo}.

\section{Demonstration}
\label{cha:demonstration}

\subsection{Data Showcase and Validation}
\label{cha:data_quality}

BuilDa can be configured to handle a wide range of parameter values, as shown in \autoref{tab:rang_building_para}. These values are provided as a guideline to illustrate possible value ranges. They represent building variations consistent with the TABULA database \cite{Loga2011-ws}, which includes buildings constructed in Germany from 1949 to the present. It is important to note that our framework is not restricted to the TABULA value ranges. Most of these values can be directly set by the user, while others are calculated indirectly through the converter layer (see \autoref{cha:Architecture}).

\begin{table}[htb]
    \small
    \caption{Example ranges for building parameter values.}
    \centering
    \setlength{\tabcolsep}{4pt} % Reduces the space between columns
    \begin{tabular}{l|c|c|c|c}
\hline
              & \textbf{Component}       & \textbf{Min} & \textbf{Max} & \textbf{BuilDa coverage}                                                                                                         \\ \hline
U-Value       & Floor, walls, roof       & 0.1          & 1.4          & directly                                                                                                                         \\
(W/m²/K)      & Window                   & 0.7          & 4.3          & directly                                                                                                                         \\ \hline
              & Floor                    & 62           & 116          & \textit{zone\_length * zone\_width}                                                                                              \\ \cline{5-5} 
Area (m²)     & Walls                    & 120          & 230          & \textit{\begin{tabular}[c]{@{}c@{}}2 * n\_floors * floor\_height * \\ (zone\_length + zone\_width) - A\_Windows\end{tabular}} \\ \cline{5-5} 
              & Roof                     & 86           & 185          & \textit{\begin{tabular}[c]{@{}c@{}}zone\_length *\\  zone\_width * f\_ARoofToAFloor\end{tabular}}                                \\ \cline{5-5} 
              & Window (e.g. south side) & 18           & 42           & \textit{\begin{tabular}[c]{@{}c@{}}floor\_height * n\_floors * \\ zone\_length * f\_AWin\_south\end{tabular}}                      \\ \hline
Heat capacity & Floor                    & 270          & 500          & directly                                                                                                                         \\
(kJ/m²/K)     & Walls                    & 50           & 660          & directly                                                                                                                         \\
              & Roof                     & 35           & 400          & directly                                                                                                                         \\ \hline
\end{tabular}
    \label{tab:rang_building_para}
\end{table}

\begin{figure}[htb]
    \centering
    \subcaptionbox{Simulation of one year.
    \label{fig:mid_temp_data_diversity}}{
        \includegraphics[width=0.48\textwidth]{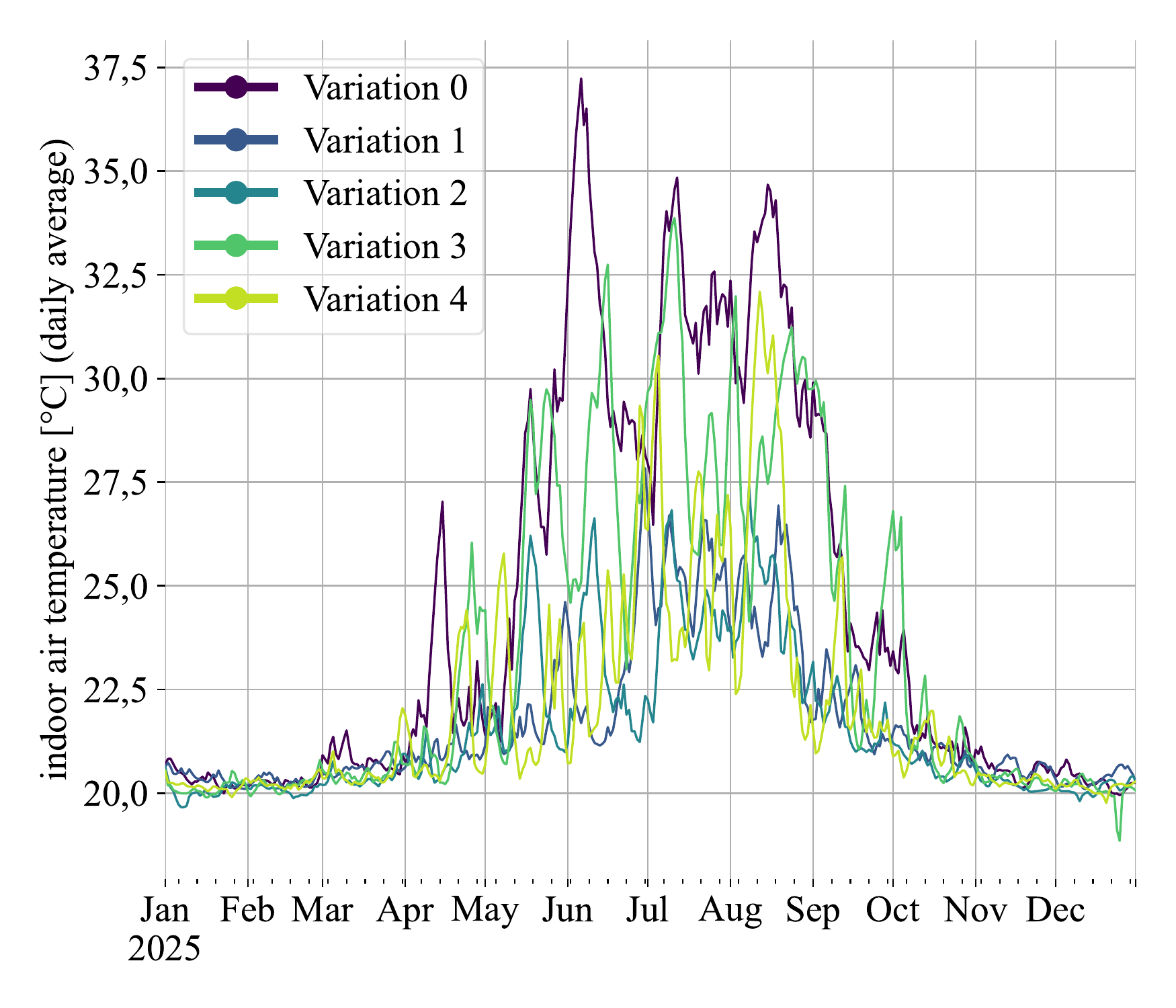}
    }
    \subcaptionbox{Simulation of one day with different controllers. Upper figure: proportional controller. Lower figure: two-point controller.\label{fig:controller_example_time_series}}{
            \includegraphics[width=0.48\textwidth]{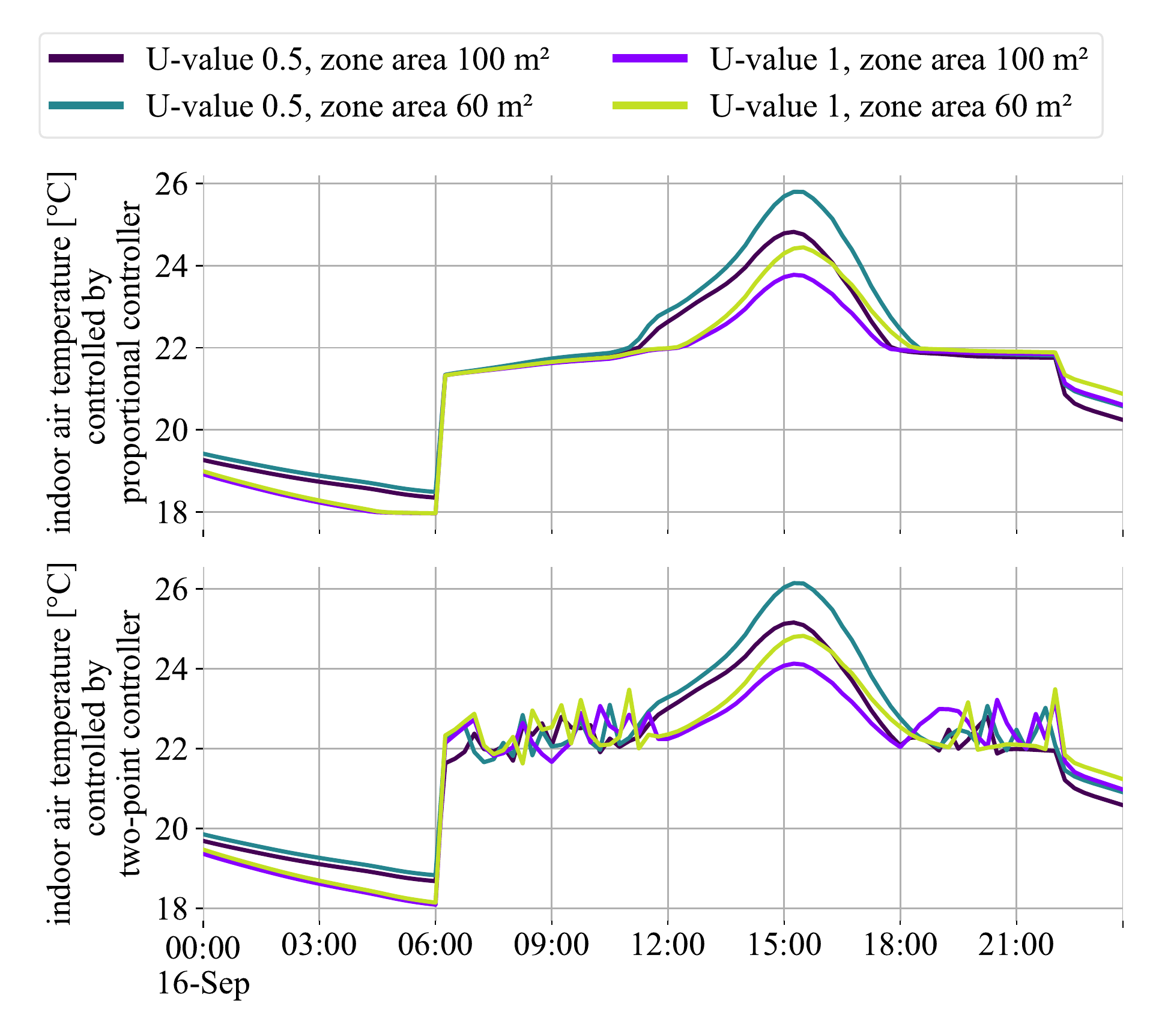}
    }
    \caption{(a) Daily average temperatures over one year for various simulations. (b) Indoor air temperatures over one day for different simulated buildings in Munich, using proportional and two-point controllers.}    
\label{fig:puzzles}
\end{figure}

\autoref{fig:mid_temp_data_diversity} depicts the daily mean zone temperatures of five different building variations over the simulation period of one year. As no cooling was applied in the simulated buildings, the mean temperatures differ greatly depending on the specific combinations of parameters of each simulation run as well as solar radiation. \autoref{fig:controller_example_time_series} shows temperature curves for four building variations on a September day, illustrating the controller's impact. The upper curves use an internal P-controller, while the lower ones use a two-point controller with night setback. Both have a 22 \textdegree C setpoint. While all buildings heat up similarly during the day, the two-point controller creates a distinct pattern by repeatedly switching the thermal source on and off to maintain the setpoint.

\begin{figure*}[!htb]
    \centering
    \includegraphics[width=0.98\linewidth]{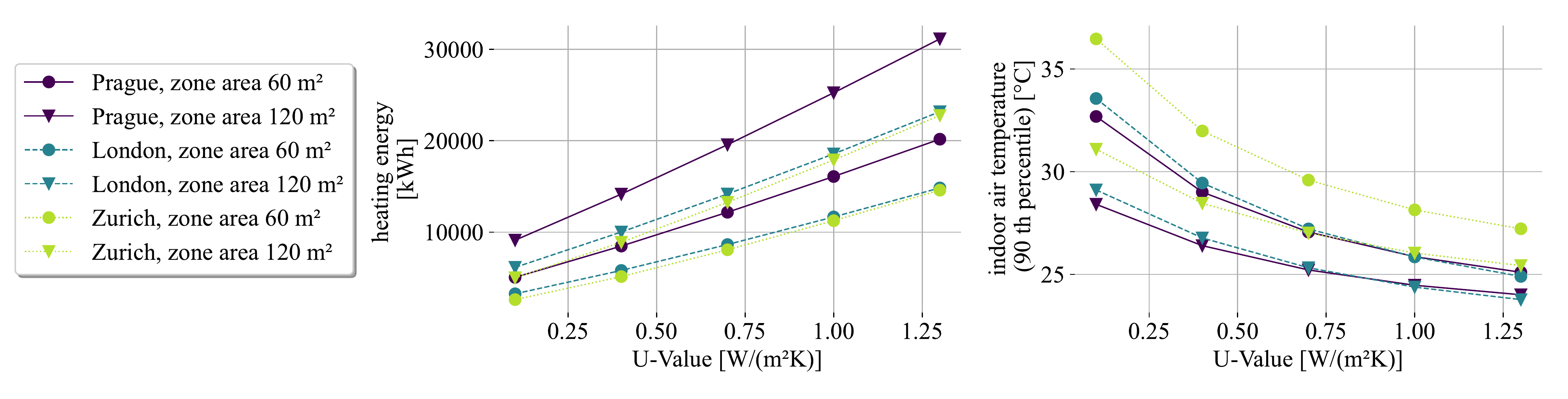}
    \caption{Relation between parameters U-value, weather, zone area and the output (left: heating energy, right: the 90th percentile of the indoor air temperature).}
    \label{fig:relation_parameter}
\end{figure*}

To further illustrate the impact of the building parameters, we show in \autoref{fig:relation_parameter} how parameters influence the energy consumption (left figure) and the 90th percentile of the indoor air temperature (right figure) of the buildings over one year. To compare different climates, we analyzed European locations with varying outdoor temperatures and solar radiation. Higher U-values increase heating energy demand while lowering indoor temperature peaks (90th percentile). Prague has the highest heating energy consumption, followed by London and Zurich, while Zurich shows the highest temperature peaks. Larger zones (120 m²) require more heating energy but tend to have slightly lower temperature peaks than smaller zones (60 m²). These results highlight the inverse relationship between heating demand and peak indoor temperatures, better insulation (lower U-value) reduces heating needs but leads to higher peak temperatures. 

To ensure accuracy of the data, we validated the model according to the ANSI/ASHRAE 140-2004 standard using the following test cases \cite{ASHRAE140}: TC600 and TC900 (fulfilled on both: annual heating and cooling), TC600FF and TC900FF (fulfilled on both: minimum, maximum and mean temperature). This provides further evidence that our data is accurate.

\subsection{Transfer Learning Example} 
\label{cha:transfer_learning}

We now present an example of how to use BuilDa for TL research by pretraining various source models and fine-tuning them to a target. We consider the problem of source building selection, as it is crucial for successful TL in building thermal dynamics to choose the right source building for knowledge transfer \cite{pinto2022sharing, 2024Similarity}. As stated in \autoref{sec:background}, investigations on the building parameter level are currently hard to execute, as accessible metadata is often missing. We therefore want to show how changes on the building parameter level of source buildings can influence TL and how BuilDa can be utilized. This study is not aimed to comprehensively explore all building parameters and their impact on TL for thermal dynamics, it is rather supposed to demonstrate how BuilDa can be used to address research gaps.

Our study design is as follows. For the source buildings, we select three different building parameters that we want to vary with BuilDa. We chose parameters that have a strong influence on the building's thermal dynamics, namely, the U-values of the walls, the heat capacity of the walls, and the floor area. The ranges we select for each parameter resemble the ranges of single-family houses depicted in \autoref{tab:rang_building_para}. This results in wall U-values 0.1, 0.7 and 1.4 W/m²K, wall heat capacities 50, 250, and 450 kJ/m²K and floor areas 60, 90, and 120 m². We employ all possible parameter combinations, getting 27 different source buildings (sr1-sr27). To identify each source, we add an appendage to its name, e.g., sr19\_$bca$. The position of the letters in the appendage encode the parameters in the order U-values (1), heat capacity (2) and floor area (3). The letters $a$,$b$,$c$ represent the parameter values, with $a$ being the lowest value and $c$ the highest. Thus, the combination $bca$ refers to the building with a U-value of 0.7 W/m²K, heat capacity of 450 kJ/m²K, and floor area of 60 m². For the target building, we chose a parameter combination on the edge of the distribution with parameter values: U-value 0.15 W/m²K, heat capacity 430 kJ/m²K, and floor area 110 m². By choosing a target on the edge of the distribution, effects on the building parameter level should become more easily apparent. We further assume a monolithic wall structure, a window-to-wall ratio of 0.14 and the weather of Munich. We use the internal controller for heating with a lower and upper setpoint of 18\textdegree C (night) and 22\textdegree C (day), respectively. Except for the varying parameters, all other building parameter values are kept the same for the source and target buildings. The task we consider is the forecast of the indoor air temperature, depending on current conditions, i.e., the weather (direct and diffuse solar radiation, outside air temperature), the indoor air temperature, and the control signal of the thermal source. For the data-driven model, we chose an LSTM \cite{LSTM1997} with an additional fully connected layer to forecast the next 4 steps (1 hour), similar to the implementation in \cite{2025gentl}. We use the ADAM \cite{kingma2014adam} optimizer for the parameter updates of the LSTM with the mean squared error cost function.

We generate one year of data for the target building and for each of the 27 source buildings. We use the source building data to pretrain 27 distinct source models. For each source model, we first perform hyperparameter tuning to get individual LSTM configurations, then we pretrain the models. Afterwards, each source model gets fine-tuned to the target. For fine-tuning, we assume only limited target data is available, i.e., 30 days in January. The rest of the year of the target data is used for testing the prediction quality of the fine-tuned models. The fine-tuned models inherit the hyperparameters from the pretrained source models. To compare the prediction performance of the fine-tuned models, we also train a target model from scratch. Training from scratch means the data-driven model is trained with the target data (30 days) without having a pretrained model as a starting point. We also performed hyperparameter tuning for the target model. We then compare the 27 fine-tuned models with the model trained from scratch.

\begin{figure}[htb]
    \includegraphics[width=0.97 \linewidth]{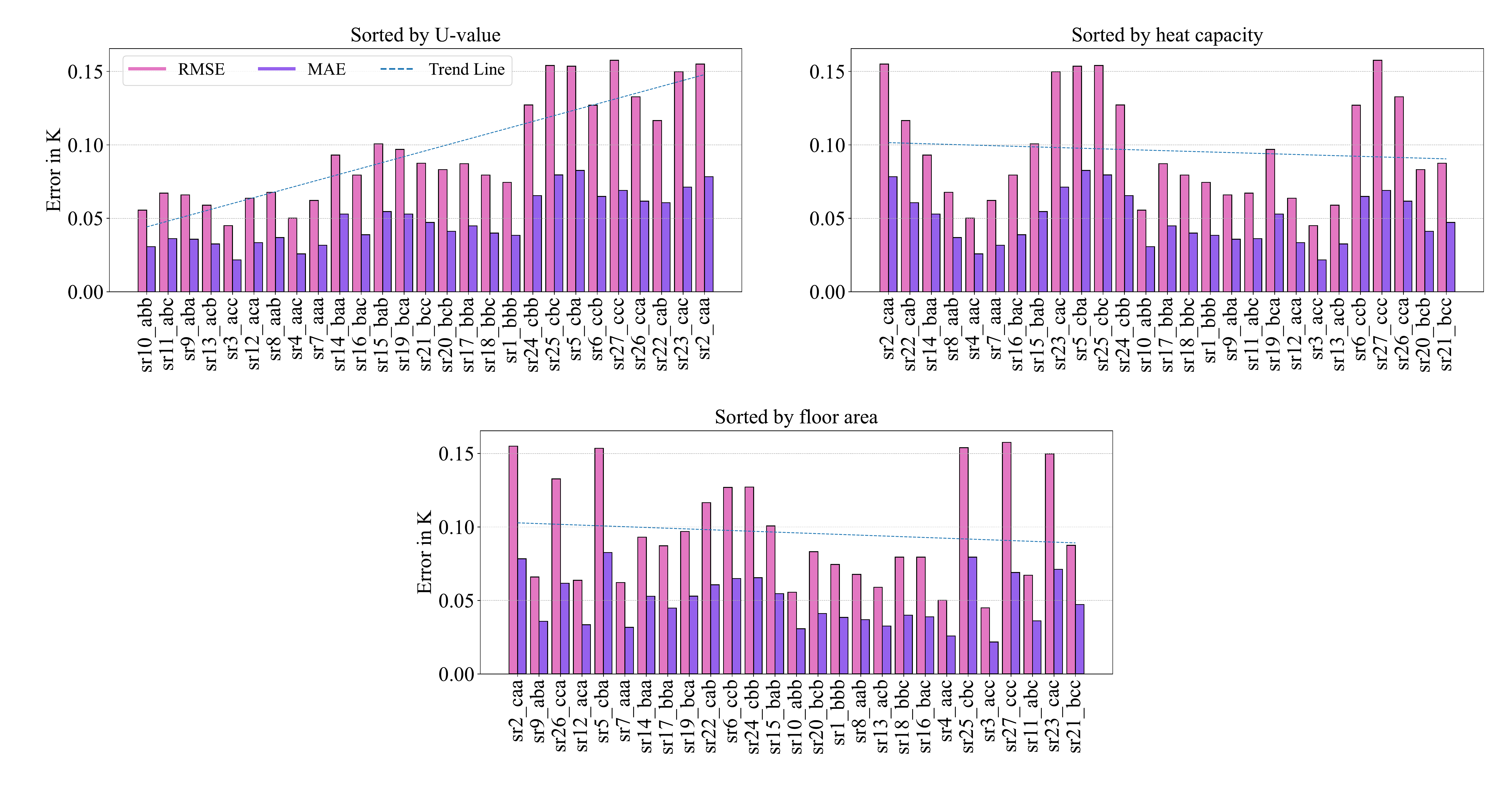}
    \centering
    \caption{Demonstration for TL. The results of the fine-tuned models are sorted with respect to the parameter values of U-Value, heat capacity and floor area.}
    \label{fig:TL}
\end{figure}

The fine-tuning results are shown in \autoref{fig:TL}. The bar plots show the mean average error (MAE) and the root mean square error (RMSE) of the fine-tuned models. The results are sorted with respect to each varied parameter with parameter values rising from left to right. For example, in the upper left bar plot the source buildings are sorted with respect to U-value, depicting the source buildings with the lowest U-values on the left and the ones with higher U-Values to the right. Additionally, we added a trend line through the RMSE. The RMSE of the model from scratch is \textbf{1.03}, the MAE is \textbf{0.74}. As can be seen in \autoref{fig:TL}, the RMSE and MAE values of the TL models are lower than the model trained from scratch, indicating superior performance of the former.
The reason is that the source and target buildings are still similar, as only a small fraction of all defining building parameters have been varied, and all buildings are exposed to the same weather. However, there are distinct differences when it comes to the selected sources, as the best-performing model sr3\_$acc$ has an RMSE of 0.05, and the worst performance is of the model sr27\_$ccc$ with an RMSE of 0.16. The first is rather unsurprising, as it is based on the source model with parameter values closest to the target. The model sr2\_$caa$, which is furthest away to the target regarding the parameter values of its source has the second-worst performance. There is also a clear performance trend visible after sorting for U-values. Source parameter values closer to the target parameter values seem to imply more similar thermal dynamics of the buildings and, thus, lower errors after fine-tuning. Yet, the influence of the heat capacity and the floor area parameter are not as clearly visible as the error trend lines lower only slightly while the source parameter values get closer to the target values.
Surprisingly, there are some parameter value combinations that do not follow the trend lines. For example, regarding floor area, the combination sr9\_$aba$, with a floor area of 60 m² performs rather well. Similarly, for heat capacity, the model sr4\_$aac$ with heat capacity of 50 kJ/m²K shows good performance while the model sr27\_$ccc$ with heat capacity of 450 kJ/m²K exhibits the worst performance. The results seem to suggest, that the influence of the U-value parameter dominates the influence of the other two parameters.
In summary, the results anecdotally demonstrate that TL models often exceed the model trained from scratch in terms of prediction accuracy when only limited training data is available. However, it is still difficult to determine which source is best suited for the target, as similar building parameters do not necessarily imply similar building thermal dynamics. 
It is unclear which parameters are most important to consider for selecting the right source for successful TL and which parameter value combinations lead to the best performance after TL. This underlines that in-depth studies on building similarity are very much needed. We want to emphasize, that in addition to the selected parameters in our short study, it is important to cover numerous other building parameter configurations, different climatic conditions, seasonality, occupancy schedules, etc., as well as different target buildings to get a clear picture. BuilDa gives researchers a data generation tool to perform such comprehensive studies.

\section{Conclusion and Future Work}
\label{cha:conclusion}

We introduced BuilDa, a highly flexible data generation framework for generating large amounts of high-fidelity thermal building dynamics time series data, designed for TL research. We described the variable physical building model in detail, presented the overall architecture of the framework and emphasized the distinct features of BuilDa. With BuilDa and its high-fidelity building model, it is easy to simulate many different building variations without expert knowledge in building simulation. Building parameters, weather scenarios, occupancy and the control of the simulation can be individually configured by the user. We showcased examples of the heterogeneous data BuilDa can generate and we demonstrated the usage of data with a short TL example. 

With BuilDa, many opportunities for TL research arise. One possible research application can be the pretraining of generalized Transformer architectures, comparable to works like \cite{Hertel2023TansLoad}, but for building thermal dynamics forecasting instead of building load forecasting. Such models can then be fine-tuned to a target building and act as a base model for control strategies, such as Model Predictive Control. Similar approaches are already used in the area of building power demand forecasting \cite{Gokhale2023TransformerDemand}. 
Further, BuilDa enables now comprehensive building similarity studies with a level of detail, which is not yet present in the TL community for building thermal dynamics. Moreover, we aim to extend the framework for reinforcement learning (RL) research, similar to systems such as Sinergym \cite{Sinergym2021}. Our current Python interface to the FMU is relatively easy to configure as a Gymnasium \cite{towers_gymnasium_2023} training environment for RL. Especially the variability of our physical model may bring additional research benefits, as many different building variations acting as RL training environments can be configured and simulated in one go. Additionally, we plan to provide other base models representing different building types and want to incorporate multi-zone buildings. This way, we hope to extend the framework to a universal, easy to use TL research platform for data-driven modeling and research for control-oriented RL.

\section*{Acknowledgments}

Special thanks to  Felix Koch and David Broos for their assistance with the project programming.

%Place the acknowledgments section, if needed, after the main text, but before any appendices and references. The section heading is not numbered. These instructions are adapted from the version by Roeder, T. M. K., Frazier, P. I., Wotton, H., Szechtman, R., Yooh, H. L., Harker, J., and Zhou, E. Those instructions were adapted from WSC instructions with permission from WSC BoD~\cite{WSC} that have been iteratively updated and improved by proceedings editors and several other individuals, who are too numerous to name separately since the first set of instructions were written by Barry Nelson for the 1991 WSC.

\appendix

\section{Appendices} \label{app:quadratic}

\begin{table}[htb]
\scriptsize\caption{Variable building simulation parameters of BuilDa.}
\centering
\scriptsize
\begin{tabular}{l|ll}
Parameter name                   & Explication       & Unit                 \\  
\hline
        zone\_length, zone\_width                    & Length (from east to west) and width (from north to south) of zone                                  & m    \\
        n\_floors, floor\_height                    & Number and height of floor levels                                              & m, -    \\ 
        fAWin\_south    & Window to wall fraction on southern,                                    & -    \\ 
         \ \ \ \ \ \ \ \textit{(analogously for west, north, east)}                      &   \ \ \ \ \ \ \ western, norther and eastern wall  & -    \\ 
        fATransToAWindow                 & Fraction of transparent window area to overall window area                      & -    \\ 
        fARoofToAFloor                   & Fraction of roof area to floor area (if inclined roof, this factor is > 1)     & -    \\ 
        fAInt                            & Factor of exterior wall surfaces to interior wall surfaces (both sides)      & -    \\ 
        
        UExt                             & U-value of the external walls                                                & W/(m²K) \\
        \ \ \ \ \ \ \ \textit{(analogously for intWall, floor, roof)}         & \ \ \ \ \ \ \ internal wall, floor and roof                         & W/(m²K)    \\

        heatCapacity\_wall               & Heat capacity of the exterior walls related to its area                                                & J/(m²K) \\ 
        \ \ \ \ \ \ \ \textit{(analogously for intWall, floor, roof)}         & \ \ \ \ \ \ \ internal wall, floor and roof                         & J/(m²K)    \\
                heatCapacity\_furniture\_per\_m2 & specific heat capacity of the furniture related to the floor area                                              & J/(kg*K*m²)                \\
        UWin                             & U-value of the windows                                                       & W/(m²K) \\ 
        thermalZone.gWin                & G-value of the windows            & -  \\

        weaDat.fileName                  & Path(s) to weather file(s)                                                     & -    \\ 
        internalGain.fileName            & Path(s) to internal gain profile file(s)                                           & -    \\ 
        hygienicalWindowOpening.fileName  & Path(s) to hygienical window opening profile file(s)                                & -    \\ 
        
        heatRecoveryRate                 & Heat recovery rate of ventilation system                                                                       & - \\
        airChangeRate                    & Air change rate due to ventilation system or infiltration                                                                          & -  \\        
        roomTempLowerSetpoint            & Lower temperature setpoint (for night setback on internal controller)      & -    \\
        roomTempUpperSetpoint            & Upper temperature setpoint (for night setback on internal controller)      & -    \\
        UseInternalController             & Decides if model internal controller for heating is used                     & -    \\ 
        extWall\_C\_distribution   & Heat capacity distribution profile for external wall,    & -    \\
        \ \ \ \ \ \ \ \textit{(analogously for intWall, floor, roof)}         & \ \ \ \ \ \ \ internal wall, floor and roof                         & -    \\
        extWall\_R\_distribution       & Heat resistance distribution profile for external wall,        & -    \\
        \ \ \ \ \ \ \ \textit{(analogously for intWall, floor, roof)}         & \ \ \ \ \ \ \ internal wall, floor and roof                         & -    \\
        internalGainsConvectiveFraction  & Fraction of internal gains that are convective                               & -    \\
        heatingConvectiveFraction         & Fraction of heating that is convective                                       & -    \\
\end{tabular}
\label{tab:allParameters}
\end{table}

% Please don't change the bibliographystyle style
\bibliographystyle{scsproc}

% AUTHOR: Include your bib file here
\bibliography{buildabib}

\section*{Author Biographies}

\textbf{\uppercase{Thomas Krug}}, M.Sc. is a Ph.D. candidate at the Technical University of Applied Sciences Rosenheim. His research interests include data-driven modeling of building dynamics, transfer learning, and reinforcement learning for intelligent building control. His email address is \email{thomas.krug@th-rosenheim.de}.

\textbf{\uppercase{Fabian Raisch}}, M.Sc. is a Ph.D. candidate at the Technical University of Munich. His research interests include transfer learning, data-driven system identification, and control of the building environment. His email address is \email{fabian.raisch@tum.de}.

\textbf{\uppercase{Dominik Aimer}}, M.Sc. is a research associate at the Technical University of Applied Sciences Rosenheim. His research interests include 
technical monitoring,
IoT and simulation development,    
physical modeling,
energy optimization,
data-driven modeling.
His email address is \email{dominik.aimer@th-rosenheim.de}.

\textbf{\uppercase{Markus Wirnsberger}}, M.Sc. is a Ph.D. candidate at the Technical University of Applied Sciences Rosenheim. His research focuses on the simulation and monitoring of building HVAC systems to optimize energy consumption and comfort. His email address  is \email{markus.wirnsberger@th-rosenheim.de}.

\textbf{\uppercase{Ferdinand Sigg}}, M.Sc. is a Ph.D. candidate at the Technical University of Applied Sciences Rosenheim. His research interests include building physics, energy efficient HVAC systems and building performance simulation. His email address is \email{ferdinand.sigg@th-rosenheim.de}.

\textbf{\uppercase{Benjamin Schäfer}}  is an Assistant Professor at the Department of Informatics at KIT, Germany, since 2023. He investigates power systems with the help of (explainable) machine learning and stochastic modeling. Prior to joining KIT, he worked in Göttingen (Germany), Dresden (Germany), London (United Kingdom), Tokyo (Japan) and Ås (Norway), graduating with a PhD thesis in 2017. His email address is \email{benjamin.schaefer@kit.edu}.

\textbf{\uppercase{Benjamin Tischler}} is a Professor at the Faculty of Applied Sciences and Humanities at the Technical University of Applied Sciences Rosenheim, Germany. His research interests include applying machine learning for modeling and optimal control of energy systems and buildings, machine learning operations and language models. Prior to joining THRO he worked in Cologne (Germany) and Munich (Germany), graduating with a PhD thesis in 2015. His email address is \email{benjamin.tischler@th-rosenheim.de}.

\end{document}